\documentclass[letterpaper, 10 pt, conference]{ieeeconf}  

\IEEEoverridecommandlockouts                              

\usepackage{microtype}
\usepackage{graphicx}
\usepackage{subfigure}
\usepackage{multicol}
\usepackage{tabularx}
\usepackage[dvipsnames]{xcolor}
\usepackage{booktabs} 
\usepackage{hyperref}

\usepackage{algorithm}

\usepackage{graphics}
\usepackage{graphicx}
\usepackage{float}
\usepackage{import}
\usepackage{pgfplots}
\usepackage{multirow}
\usepackage{amssymb}

\usepackage[tableposition=top]{caption}
\usepackage{booktabs}
\usepackage{siunitx}
\usepackage{amsmath}
\numberwithin{equation}{section}
\newtheorem{theorem}{Theorem}[section]
\newtheorem{assumption}{Assumption}[section]
\newtheorem{lemma}{Lemma}[section]
\renewcommand{\theequation}{\arabic{section}.\arabic{equation}}
\definecolor{color1}{rgb}{0,0.266666666666667,0.105882352941176}

\title{\LARGE \bf
Off Environment Evaluation Using Convex Risk Minimization}

\author{Pulkit Katdare, Shuijing Liu, and Katherine Driggs Campbell
\thanks{The authors are with the department of Electrical and Computer Engineering, University of Illinois at Urbana-Champaign, Champaign, IL, 61820 \texttt{\{katdare2, sliu105, krdc\}@illinois.edu }}
} 

\DeclareUnicodeCharacter{2212}{-}
\begin{document}

\maketitle
\thispagestyle{empty}
\pagestyle{empty}

\begin{abstract}

Applying reinforcement learning (RL) methods on robots typically involves training a policy in simulation and deploying it on a robot in the real world. Because of the model mismatch between the real world and the simulator, RL agents deployed in this manner tend to perform suboptimally. To tackle this problem, researchers have developed robust policy learning algorithms that rely on synthetic noise disturbances. However, such methods do not guarantee performance in the target environment. We propose a convex risk minimization algorithm to estimate the model mismatch between the simulator and the target domain using trajectory data from both environments. We show that this estimator can be used along with the simulator to evaluate performance of an RL agents in the target domain, effectively bridging the gap between these two environments. We also show that the convergence rate of our estimator to be of the order of ${n^{-1/4}}$, where $n$ is the number of training samples. In simulation, we demonstrate how our method effectively approximates and evaluates performance on Gridworld, Cartpole, and Reacher environments on a range of policies. We also show that the our method is able to estimate performance of a 7 DOF robotic arm using the simulator and remotely collected data from the robot in the real world. 
\end{abstract}
\section{Introduction}
Reinforcement learning (RL) has demonstrated success in a wide variety of applications such as AlphaGo~\cite{DBLP:journals/nature/SilverSSAHGHBLB17},  recommendor systems~\cite{DBLP:journals/corr/abs-1810-12027}, and robotics~\cite{hoel2019combining,chang2020sound,DBLP:conf/icml/HaarnojaZAL18,DBLP:journals/corr/LillicrapHPHETS15}, albiet with some concerns regarding their sensitivity to hyperparameters~\cite{DBLP:conf/aaai/0002IBPPM18}. Using such hyperparameter sensitive algorithms might lead to an RL agent learning sub-optimal policies, which gets further exacerbated when there is a mismatch between the training environment and the intended target environment. 

A typical approach to robot learning involves training and optimization of policy in a software based simulator of the same robot~\cite{levine2016end, DBLP:journals/corr/AndrychowiczWRS17}. A robot trained in such a manner tends to perform sub-optimally in the intended target environment~\cite{DBLP:conf/icra/PengAZA18}. The primary reason behind this sub-optimal performance that these simulators fail to capture the nuances and the real world environment accurately, which is often referred to as the Sim2Real gap~\cite{DBLP:conf/wafr/TzengDHFALSD16}. These nuances either require strong assumptions (e.g., modelling contact forces) or are extremely difficult to model (e.g., human-robot interaction). The resulting sub-optimality is not only hard to characterize but it also might lead to potentially hazardous scenarios in high risk environments~\cite{li2018learning}.

To account for such model mismatches, Robust RL algorithms learn a policy which is robust to synthetic perturbations in the simulator~\cite{DBLP:conf/iros/TobinFRSZA17, DBLP:journals/corr/TzengDHFPLSD15}. These synthetic perturbations are created under the assumption that the discrepancy between the simulator and the target domain is due to an misidentification of simulator parameters. These simulators are often governed by a model with a lower complexity than the one in the target domain or struggle to capture real-world disturbances~\cite{ma2018improved}. Applying robust RL techniques to account for significant modeling error or higher complexity altogether will likely not be enough to ensure a successful domain transfer without extensive real world validation~\cite{DBLP:journals/corr/abs-2012-03806}. We believe that one of the reasons such policies fail to work in practice is due to a fundamental misunderstanding and lack of formalism of model mismatch and policy evaluation across model or environment changes. For example, using robot learning to manipulate and solve a Rubik's cube required more than two years of research on developing a state-of-the-art robust learning algorithms as compared to six months it took them to learn the same task in simulation~\cite{DBLP:journals/corr/abs-1910-07113}.

We begin our work with an assumption that it is possible to collect data of agent behaviors from the target environment. In this paper, we use a small amount of data from the target domain to learn a mismatch function that modifies simulator performance to be more reflective of true performance in deployment. We quantify this mismatch function as the as a transition ratio estimation problem between the simulator and the target domain, both of which are modelled as Markov Decision Processes (MDP). In order to estimate this transition ratio, we further pose this estimation as an optimizer of a convex risk minimization algorithm that utilizes trajectory samples from both the target and the simulator environments. We then use this estimated ratio along with the simulator to evaluate a robot agent's performance in the target domain. 

We demonstrate that our estimator tends to approximate the true transition ratio with the accuracy error decreasing with the rate of $n^{-1/4}$, where $n$ is the number of transition samples used to estimate the ratio. We also illustrate the effectiveness of our algorithm to approximate the returns for a short time horizon by proving an exponential dependence of our estimator on the time horizon.

We present the following contributions:
(1) We introduce off-environment evaluation, which learns to incorporate model mismatch into the evaluation of a fixed policy across different environments; 
(2)  We use results from convex risk minimization to efficiently learn the model mismatch as the ratio of transition probabilities across two environments; 
(3) Using tools from statistical learning theory, we bound the error rate of our estimator as a function of the number of samples from the target domain; and 
(4) We demonstrate the utility of our approach on three task settings, a large discrete example (gridworld), and continuous environments (cartpole, reacher and kinova).

This paper is organized as follows. Section \ref{sec:litreview} reviews relevant literature. Section \ref{sec:prelim} introduces necessary preliminaries to pose our problem. Section \ref{sec: OEE} discusses our proposed method of off-environment policy evaluation. Section \ref{sec: error} proves the theoretical error bounds of our algorithm. We demonstrate our results by starting with a discrete-state, discrete-action MDP and building up to a real robot manipulator in Section \ref{sec: expt} and conclude our paper in Section \ref{sec: conclude}. 
\section{Related Work}
\label{sec:litreview}
In this section, we review relevant literature that considers model mismatch, covariate shift, and off-policy evaluation  which inspired our formulation of off-environment evaluation.
\subsection{Model Mismatch and Covariate Shift}
Model mismatch between two domains is well studied in machine learning, often in context of estimating covariate shift~\cite{ DBLP:journals/jmlr/AgarwalLS11, DBLP:journals/jmlr/SugiyamaKM07}. Covariate shift occurs when the training and the test data are sampled from different distributions. The goal of the above work is to learn a classifier that generalizes well on the test distribution while being learned entirely on the training dataset. 

The impact of covariate shift is often mitigated by estimating the distributional mismatch factor between the training and the testing distributions $\beta(x)$~\cite{conf/aaai/ReddiPS15}. The factor is then used to re-weight samples on the training dataset such that higher weights are assigned to data points which have a higher likelihood of being in the test distribution than the training distribution. 
Our work extends this idea of using a distributional mismatch factor to estimate mismatch between two Markov Decision Processes (MDP), not just distributions. 
\subsection{Off-Policy Evaluation}
Approximating performance by a target policy given access to data from a different behavior policy falls under the class of Off Policy Evaluation (OPE) in reinforcement learning~\cite{DBLP:journals/corr/abs-1911-06854}. OPE literature is generally divided into three class of methods, inverse propensity scoring~\cite{Precup00eligibilitytraces, DBLP:journals/corr/abs-1806-01347}, direct methods ~\cite{pmlr-v97-le19a, DBLP:journals/corr/abs-1802-03493}, and hybrid methods ~\cite{pmlr-v48-thomasa16,DBLP:journals/corr/JiangL15}. 

We are inspired by a subset of direct methods that aims to minimize a convex risk criteria to estimate the average state-action density ratios between the behavior and the target policy based on the data from behavior policy and the functional form of target policy returns~\cite{DBLP:conf/nips/NachumCD019}. The density ratio is later used to re-weight the reward function to estimate returns using data available from the behavior policy. Similar to the direct set of methods, our work takes advantage of a convex risk minimization algorithm to estimate density ratios between the data but unlike these set of methods we use convex risk minimization to estimate the transition ratio between two different MDP environments. 

\section{Preliminaries} \label{sec:prelim}

In this section, we introduce our notation for MDPs and reinforcement learning and present the preliminaries for importance sampling and the associated estimators.

\subsection{MDPs and Reinforcement Learning}

RL problems are modelled as a Markov Decision Process (MDP), which are represented as a 5-tuple $\mathcal{M} = \langle\mathcal{S}, \mathcal{A}, \mathcal{P}, r, S_0 \rangle$, where $\mathcal{S}$ is the state space, $\mathcal{A}$ is the set of available actions to the agent, $\mathcal{P}$ gives us the underlying transition probability of the next state given the current state and action taken in the environment, and $S_0$ is the distribution from which initial states $s_0$ are sampled. 
We define a stationary policy $\pi: \mathcal{S} \rightarrow \Delta(\mathcal{A})$ as a probability distribution over actions for each states. 
At every time step $t$, the environment evolves to state $s_{t+1} \sim \mathcal{P}(\cdot| s_t, a_t)$ from state $s_{t}$ by taking an action $a_t$ according to a policy $\pi(\cdot| s_t)$. We define value of a policy by the function $J_{\mathcal{P}}(\pi)$ which calculates the average cumulative discounted sum of rewards $r: \mathcal{S} \times \mathcal{A} \rightarrow \mathbb{R}$ over an infinite time horizon:
\begin{align}
\begin{split}
    & J_{\mathcal{P}}(\pi) = \mathbb{E}\left[\sum_{t=0}^\infty \gamma^t r(s_t, a_t)\right] \\
     & s_0 \sim S_0 \quad \forall t, a_t \sim \pi(.|s_t), s_{t+1} \sim \mathcal{P}(.|s_t, a_t) 
     \end{split}
\end{align}
where $\gamma \in (0, 1]$ is the discount factor. 
An alternative formulation to calculate $J_{\mathcal{P}}(\pi) = \frac{1}{1 - \gamma}\mathbb{E}_{(s, a) \sim d^{\pi}}[r(s, a)]$ involves calculating expectation under the distribution $d^{\pi}_{\mathcal{P}}(s, a)$, which is the normalized discounted stationary distribution: 
\begin{align}\label{eq: stationarydist}
\begin{split}
    d_{\mathcal{P}}^\pi(s, a) &= (1-\gamma) \sum_{t=0}^\infty \gamma^t \mathcal{P}r(s_t = s, a_t = a) \\
    &  s_0 \sim S_0, \forall t, a_t \sim \pi(\cdot|s_t), s_{t+1} \sim \mathcal{P}(\cdot|s_t, a_t). 
    \end{split}
\end{align}
We similarly define the normalized discounted state-action-state distribution. We observe that the relationship between $d_{\mathcal{P}}(s, a)$  and $d_{\mathcal{P}}(s, a, s')$ can be expressed by:
\begin{align}\label{eq: statedistribution}
    d_{\mathcal{P}}^\pi(s, a, s') = d_{\mathcal{P}}^\pi(s, a) \mathcal{P}(s'|s, a).
\end{align} 
It is worth noting that this relationship becomes necessary to estimate the mismatch between two MDPs. To estimate this mismatch, we first define two MDP environments: training $MDP_{tr}$ and testing $MDP_{te}$. \newline 
\begin{align}\label{eq: MDPS}
\begin{split}
    &MDP_{tr} = \left \langle \mathcal{S}, \mathcal{A}, \mathcal{P}_{tr}, r, S_0 \right\rangle, \\
    &MDP_{te} = \left \langle \mathcal{S}, \mathcal{A}, \mathcal{P}_{te}, r, S_0 \right \rangle
    \end{split}
\end{align}
Note that both these MDPs are identical in all elements except the environment model. 

Mathematically, we wish to estimate $J_{\mathcal{P}_{te}}(\pi)$ while having full access to $MDP_{tr}$ and limited data from $MDP_{te}$. First, we wish to estimate the ratio of transition probability $\zeta = \frac{\mathcal{P}_{te}(s' | s, a)}{\mathcal{P}_{tr}(s'|s, a)}$ and use this ratio to estimate $J_{\mathcal{P}_{te}}(\pi)$ using $MDP_{tr}$ and concepts from importance sampling.

\subsection{Importance Sampling}
Importance sampling is a technique to estimate the statistical properties of a given probability distribution, given access to samples from a similar distribution. Consider two probability distributions on the same sample space $(\Omega, \mathcal{F}, P)$ and $(\Omega, \mathcal{F}, Q)$.  We want to infer a statistical property $\mathbb{E}_{x \sim P}[G(x)]$, for any $G: \Omega \rightarrow \mathbb{R}$, with only access to samples from $x \sim Q$. Under the assumption that $\forall x \in \Omega$, if $P(x) > 0 \Rightarrow Q(x) > 0$, we can re-write the expectation in the following form:
\begin{align}
\begin{split}
    E_{x \sim P}[G(x)] &= \sum_{x \in \Omega} P(x) G(x) \\
    &= E_{x \sim Q}\left[\frac{P(x)}{Q(x)} G(x)\right] \label{eq: impsamp}
    \end{split}
\end{align}
The term $\frac{P(x)}{Q(x)}$ is often referred to as the importance sampling factor $\beta(x)$. When using inverse propensity scoring methods for off policy evaluation methods, this ratio $\beta(x)$ is generally known in advance. However, in some applications, this ratio needs to be estimated based on samples from the distribution $x \sim P$ and $x' \sim Q$. In the next section, we discuss about one such method to estimate the ratio $\beta(x)$. 

\subsection{Estimating the Importance Sampling Factor} \label{sec: is}
In our approach, we wish to learn $\beta(x)$ as a function of the state space. For that purpose, we look at the entropy regularisation class of methods which re-formulates the f-divergence between these two distributions in terms of their convex dual~\cite{DBLP:journals/tit/NguyenWJ10}. Specifically, the Kullback-Leiber divergence between two distributions $P$ and $Q$ can be rewritten as a maximisation problem over a space of functions $\mathcal{G}$:
\begin{align}\label{eq: expoptimizer}
    \hat{\mathcal{D}}_{KL}(P || Q) =  \max_{g \in \mathcal{G}}\sum_{x' \in \Omega}& P(x') \log(g(x'))  \\
    &- \sum_{x \in \Omega} Q(x) g(x)  +  1 \nonumber.
\end{align}
Note that the optima of the above problem is $g^* \in \mathcal{G} = \frac{P(x)}{Q(x)}$, as desired. In practice, we only have access to $n$ samples each from $P$ and $Q$. Thus, the optimal $g^*$ is approximated by searching for $\hat{g}_n$ that minimizes the empirical loss along with a regularizer $I(g)$ to prevent over-fitting:
\begin{align}\label{eq: optimizer}
    \hat{g}_n = \arg \min_{g \in \mathcal{G}}
    \frac{1}{n}\sum_{i=1}^n g(x_i') - \frac{1}{n}\sum_{i=1}^n \log(g(x_i)) + \frac{\lambda_n}{2} I(g)^2
\end{align}
In this work, we use neural networks as the class of functions $\mathcal{G}$ to show the effectiveness of our approach. In practice, researchers have also used other different function spaces like Reproducible Kernel Hilbert Space (RKHS)~\cite{DBLP:journals/corr/abs-1810-12429}. For small scale (finite state, finite action) experiments, the function space is often represented by matrices~\cite{DBLP:conf/nips/NachumCD019}. 

\section{Off Environment Policy Evaluation}\label{sec: OEE} 
We present an algorithm to evaluate performance of an agent in the testing MDP using a training MDP along with minimal data from the testing MDP. We first show that to approximate returns using the training MDP, one requires the ratio of transition probabilities of the two distributions $\zeta$. We further use the convex optimization method to propose an algorithm to estimate this ratio. 

\subsection{Methodology}
Consider a training and a testing MDP (equation \ref{eq: MDPS}). Given only a few empirical samples from the testing MDP $\{s_t, a_t, s_{t+1}\}^{tr}_{t=1, ... ,N}$, we wish to estimate $J_{\mathcal{P}_{te}}(\pi)$ using access to environment $MDP_{tr}$. To do this, consider the following expression for $J_{\mathcal{P}_{te}}(\pi)$:
\begin{flalign}\label{eq: optima}
\begin{split}
    J_{\mathcal{P}_{te}}(\pi) &=  \mathbb{E}\left [\sum_{t=0}^{\infty} \gamma^t r(s_t, a_t)\right] \\
    = & \mathbb{E}_{s_0 \sim S_0} [ r(s_0, a_0) ...   \\
    & \quad + \mathbb{E}_{s_1 \sim \mathcal{P}_{te}(\cdot|s_0, a_0)}[\gamma r(s_1, a_1) ...
    \\ & \quad \quad +  \mathbb{E}_{s_2 \sim \mathcal{P}_{te}(\cdot|s_1, a_1)} [\gamma^2 r(s_2, a_2)...]] ]  
    \end{split}
\end{flalign}
where $s_{t+1} \sim \mathcal{P}_{te}(\cdot|s_t, a_t)$, $\forall t =0, 1, .., \infty$. The last equation comes from the tower property of the expectation. Next, we use importance sampling equation \ref{eq: impsamp} to re-write the expectation in terms of $MDP_{tr}$:
\begin{flalign}\label{eq: value_approx}
\begin{split}
  & J_{\mathcal{P}_{te}}(\pi) =  \sum_{t=0}^\infty \mathbb{E}_{s_0 \sim S_0}\\ 
  &\prod_{k=0}^{t} \mathbb{E}_{s_{k} \sim \mathcal{P}_{tr}(\cdot | s_{k-1}, a_{k-1})} \frac{\mathcal{P}_{te}(s_k | s_{k-1}, a_{k-1})}{\mathcal{P}_{tr}(s_k | s_{k-1}, a_{k-1})}\gamma^t r(s_t, a_t)   \\
  = &\prod_{k=0}^{t} \mathbb{E}_{s_{k} \sim \mathcal{P}_{tr}(\cdot | s_{k-1}, a_{k-1})}  \zeta(s_k|s_{k-1}, a_{k-1})\gamma^t r(s_t, a_t)
  \end{split}
\end{flalign}
We thus define $\zeta$ as the ratio of conditional probabilities:
\begin{align}\label{eq: zeta}
\begin{split}
    \zeta(s_k|s_{k-1}, a_{k-1}) &= \frac{\mathcal{P}_{te}(s_k | s_{k-1}, a_{k-1})}{\mathcal{P}_{tr}(s_k | s_{k-1}, a_{k-1})} 
    \end{split}
\end{align}
Using importance sampling we thus show our methodology that can assess the policy performance in the testing environment using the training environment and the ratio of transition probabilities between two MDPs.

\subsection{Estimating $\zeta$}
We showed how a convex risk minimization algorithm can estimate the ratio of two probability distributions. Now, we use these principles to estimate the ratio of two conditional probabilities (i.e., the transition probabilities for environment models). 
Using the trajectory data collected on the testing MDP, we apply convex risk minimization algorithm previously described. We wish to calculate two types of density ratios: $\frac{d^{\mathcal{D}_{te}}(s, a, s')}{d^{\mathcal{D}_{tr}}(s, a, s')}$ and $\frac{d^{\mathcal{D}_{te}}(s, a)}{d^{\mathcal{D}_{tr}}(s, a)}$. 
We calculate $\zeta$ by calculating the ratio of these two density ratios which follows simply from the relation as mentioned in equation $\ref{eq: statedistribution}$:
\begin{align}\label{eq: ratiocond}
    \zeta(s' | s, a) = \Large \left({\frac{d^{\mathcal{D}_{te}}(s, a, s')}{d^{\mathcal{D}_{tr}}(s, a, s')}}\Large\right)\bigg/ \large \left({\frac{d^{\mathcal{D}_{te}}(s, a)}{d^{\mathcal{D}_{tr}}(s, a)}} \large \right)
\end{align}
Using the $\zeta$ estimator along with with equation \ref{eq: value_approx}, we can now approximate $J_{\mathcal{P}_{te}}(\pi)$ using just the training MDP. Note that whenever we use an estimate $\hat{\zeta}_n$ to approximate $J_{\mathcal{P}_{te}}$ we use the notation $J^{\hat{\zeta}_n}_{\mathcal{P}_{te}}$.

\section{Error Analysis}\label{sec: error}
To characterize the error of our estimator, we use existing results developed for empirical risk minimization algorithms to upper-bound the error on the KL-divergence. Using first order Taylor expansion, we later use that upper-bound the error in estimating $\zeta$. We also use this estimator error to approximate the upper-bounds in evaluating $J_{\mathcal{P}_{te}}(\pi)$ using the training environment. 
\subsection{Estimation Error}
Before analysing estimator error, we make the following assumptions.
\begin{assumption}\label{ass: bounds}
For any $g \in \mathcal{G}$, we assume that $\nu \leq g(x) \leq \mu \quad \forall x \in \Omega$. Note that $\nu < 1$ and $\mu > 1$
\end{assumption}
Note that estimating $\zeta$ (equation \ref{eq: ratiocond}) requires calculating a ratio of two entities. The lower bound assumption is to make sure we do not run into numerical issues. 
\begin{assumption}\label{ass: existence}
$\exists$ $g^* \in \mathcal{G}$ such that $g^*(x) = \frac{P(x)}{Q(x)}$
\end{assumption}
For the sake of simplicity, we will also assume that the optimizer contains the true minimizer $g^*$. Under these assumptions, we analyse the error between the empirically estimated optimizer $\hat{g}_n$ and the true estimator $g^*$.\newline 
\begin{theorem} \label{thm: estimation}
Under assumption \ref{ass: bounds}, we can upper bound the estimator error between $\hat{g}_n$ and $g^*$, with probability at-least $1- \delta$ for a sufficiently large $n$ by:
\begin{flalign}\label{eq: upperbound}
\begin{split}
&  \left \| \hat{g}_n - g^* \right\|_\infty^2 \leq Ke^{D_{\infty}(P \| Q)} \\
& \left(\sqrt{\frac{1}{n}} (\mu  + \max(\log \mu, -\log \nu)) + \sqrt{\frac{2 \log\frac{1}{\delta}}{n}}  \right)
\end{split}
\end{flalign}
For the sake of simplicity, we denote the upper-bound as a function of the function class $\mathcal{G}$ and $\delta$ as $M(\mathcal{G}, \delta)$. The detailed proofs along with additional experiments can be found in the supplementary materials\footnote{ Supplementary Materials: \href{https://pulkitkatdare.web.illinois.edu/icrasupp.pdf}{https://pulkitkatdare.web.illinois.edu/icrasupp.pdf}}\footnote{Code implementation: \href{https://github.com/pulkitkatdare/offenveval}{https://github.com/pulkitkatdare/offenveval}}. 
\end{theorem}
We observe that the error decreases at the rate of $O(n^{-\frac{1}{4}})$ and scales exponentially with the $\infty$-renyi divergence. Note that in theorem \ref{thm: estimation}, we bounded the error in estimating \ref{eq: optimizer}. Using this estimator error, we now upper-bound the error in estimating $\hat{\zeta}_n$.
\begin{lemma}\label{le: zetaerror}
Applying theorem \ref{thm: estimation}, we can bound estimation error for $\hat{\zeta}_n$ as follows:
\begin{align}
    \| \hat{\zeta}_n - \zeta^* \| &\leq \frac{\mu(1 + \nu \mu)}{\nu^2} \frac{M(\mathcal{G}, \delta)}{n^{1/4}}\\ \nonumber
\end{align}
\end{lemma}
\begin{theorem}\label{th: evalerror}
Under assumptions \ref{ass: existence} and \ref{ass: bounds} and the assumption that the reward function is bounded by a constant $\|r(s,a)\|_\infty \leq R$, we can bound the error in performance predicted by our algorithm $J_{\mathcal{P}_{te}}^{\hat{\zeta}_n}(\pi)$ for a time horizon $T$ by the following expression:\\
\begin{align}
\begin{split}
    &\|J_{\mathcal{P}_{te}}(\pi)  - J_{\mathcal{P}_{te}}^{\hat{\zeta}_n}(\pi)\|_\infty ^2 \\
    &\leq \frac{TM (\mathcal{G}, \delta)^2 R^2 \gamma }{\nu \sqrt{n}} \left(\frac{1 - (T+1)\frac{\gamma}{\nu}^T + T\frac{\gamma}{\nu}^{T+1}}{(1 - \frac{\gamma}{\nu})^2} \right) 
    \end{split}
\end{align}
\end{theorem}
with probability at-least $1-\delta$.

 We observe that although the bound above converges at the rate of $n^{-1/2}$, the bound also has an exponential dependence on the time horizon. Interestingly, the variance for inverse propensity scoring also depends exponentially on time horizon~\cite{DBLP:journals/corr/abs-1911-06854}. 
To reduce the error in practice, we employ to shorter time horizons $T \simeq 100$.
\begin{algorithm}
\textbf{Given: } $d^{\mathcal{D}}_{\mathcal{P}_{te}}$, $d^\pi_{\mathcal{P}_{tr}} $  \\
\textbf{Initialize: } $\{g_{\theta_{sas'}}, g_{\theta_{sa}}\}$ \\
\textbf{For} $t = 1,..., \text{numIterations}$ \textbf{do:} \;\\
\textbf{1: }Sample N transitions: $(s_t, a_t) \sim  d^{\mathcal{D}}_{\mathcal{P}_{te}}$, $(s_t', a_t') \sim  d^{\mathcal{\pi}}_{\mathcal{P}_{tr}}$\\
\textbf{2: }$\theta_{sa} = \theta_{sa} - \eta \nabla_{\theta_{sa}} l(g_{\theta_{sa}})$ \quad (equation \ref{eq: optimizer})\\
\textbf{3: }Sample N transitions: $(s_t, a_t, s_{t+1}) \sim  d^{\mathcal{D}}_{\mathcal{P}_{te}}$, $(s_t, a_t, s_{t+1}') \sim  d^{\mathcal{\pi}}_{\mathcal{P}_{tr}}$\\
\textbf{4: }$\theta_{sas'} = \theta_{sas'} - \eta \nabla_{\theta_{sas'}} l(g_{\theta_{sas'}})$ \quad (equation \ref{eq: optimizer})\\
\textbf{End For:}\\
\textbf{Return:} $\{\theta_{sa}, \theta_{sas'}\}$
\caption{Off Environment Evaluation}
\label{algo1}
\end{algorithm}
\section{Experiments and Results}\label{sec: expt}
Our experiments are guided by the following questions:
\begin{enumerate}
    \item How well does the ratio of two estimated importance sampling (equation \ref{eq: ratiocond}) factors work to estimate conditional probabilities? 
    \item Can we use our estimator to validate performance? What is the quality of this validation? 
    \item Can our estimator be applied to real world robots?
\end{enumerate}
We report the average performance by training 10 of these networks, all of which are initialized with different random seeds. Our algorithm is summarized in algorithm \ref{algo1}.

\subsection{Gridworld}\label{sec: gridworld}
\begin{figure*}
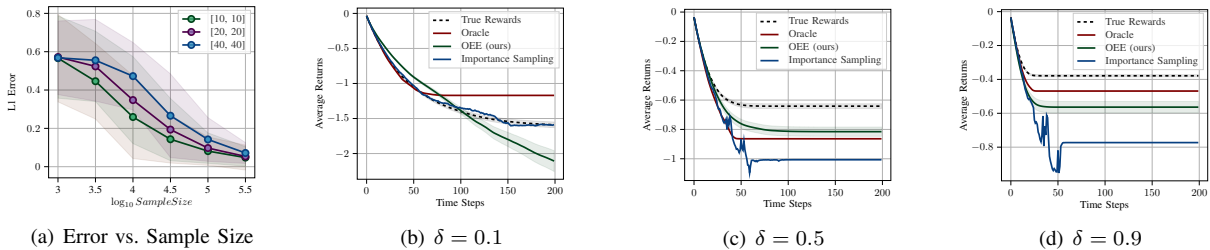

\centering
\begin{minipage}{0.23\textwidth}
\subfigure[Error vs. Sample Size]{\label{fig:gridworld_1}\scalebox{0.4}{\hspace{-1.0em}\hbox{\hspace{-1.0em}
\begin{tikzpicture}

\definecolor{color0}{rgb}{0.4,0.145098039215686,0.0235294117647059}

\definecolor{color1}{rgb}{0,0.266666666666667,0.105882352941176}
\definecolor{color2}{rgb}{0.301960784313725,0,0.294117647058824}
\definecolor{color3}{rgb}{0.0313725490196078,0.250980392156863,0.505882352941176}
\definecolor{color4}{rgb}{0.9254, 0.439, 0.0784}
\definecolor{color5}{rgb}{0.2549, 0.6823, 0.4627}
\definecolor{color6}{rgb}{0.5490, 0.4196, 0.6941}
\definecolor{color7}{rgb}{0.2117, 0.5647, 0.7529}

\begin{axis}[
legend cell align={left},
legend style={fill opacity=0.8, draw opacity=1, text opacity=1, draw=white!80!black},
tick align=outside,
tick pos=left,
x grid style={white!69.0196078431373!black},
xlabel={$\log_{10} SampleSize$},
xmajorgrids,
xmin=-0.25, xmax=5.25,
xtick style={color=black},
xtick={0,1,2,3,4,5},
xticklabels={3,3.5,4,4.5,5,5.5},
y grid style={white!69.0196078431373!black},
ylabel={L1 Error},
ymajorgrids,
ymin=-0.0581429126243319, ymax=0.835067349167269,
ytick style={color=black}
]
\path [draw=color0, fill=color0, opacity=0.1]
(axis cs:0,0.794466882722196)
--(axis cs:0,0.338283156820121)
--(axis cs:1,0.248682307125322)
--(axis cs:2,0.042599817411559)
--(axis cs:3,0.0189740310364731)
--(axis cs:4,0.00712532151648539)
--(axis cs:5,-0.0175424461792591)
--(axis cs:5,0.112754886102501)
--(axis cs:5,0.112754886102501)
--(axis cs:4,0.155594038975074)
--(axis cs:3,0.267333275058546)
--(axis cs:2,0.475871286026975)
--(axis cs:1,0.643311420156725)
--(axis cs:0,0.794466882722196)
--cycle;

\path [draw=color1, fill=color1, opacity=0.1]
(axis cs:0,0.786929428282559)
--(axis cs:0,0.357378228180006)
--(axis cs:1,0.339884261809072)
--(axis cs:2,0.120635142527828)
--(axis cs:3,0.0312104619840233)
--(axis cs:4,0.0178748801418779)
--(axis cs:5,0.00256915136679958)
--(axis cs:5,0.104424117413443)
--(axis cs:5,0.104424117413443)
--(axis cs:4,0.176560836750644)
--(axis cs:3,0.356000329572637)
--(axis cs:2,0.574548387049576)
--(axis cs:1,0.706964708282494)
--(axis cs:0,0.786929428282559)
--cycle;

\path [draw=color2, fill=color2, opacity=0.1]
(axis cs:0,0.760261248348314)
--(axis cs:0,0.376201680589005)
--(axis cs:1,0.342024152986634)
--(axis cs:2,0.296631309511059)
--(axis cs:3,0.0477046163787521)
--(axis cs:4,0.0280280200259151)
--(axis cs:5,0.0172447279885872)
--(axis cs:5,0.125850572685084)
--(axis cs:5,0.125850572685084)
--(axis cs:4,0.256236505601984)
--(axis cs:3,0.484193682897639)
--(axis cs:2,0.648388824762504)
--(axis cs:1,0.768332020732766)
--(axis cs:0,0.760261248348314)
--cycle;

\addplot [ultra thick, color1, mark=*, mark size=3, mark options={solid,fill=color5}]
table {%
0 0.566375019771159
1 0.445996863641024
2 0.259235551719267
3 0.143153653047509
4 0.0813596802457794
5 0.0476062199616209
};
\addlegendentry{[10, 10]}
\addplot [ultra thick, color2, mark=*, mark size=3, mark options={solid,fill=color6}]
table {%
0 0.572153828231282
1 0.523424485045783
2 0.347591764788702
3 0.19360539577833
4 0.0972178584462609
5 0.0534966343901213
};
\addlegendentry{[20, 20]}
\addplot [ultra thick, color3, mark=*, mark size=3, mark options={solid,fill=color7}]
table {%
0 0.56823146446866
1 0.5551780868597
2 0.472510067136781
3 0.265949149638195
4 0.14213226281395
5 0.0715476503368358
};
\addlegendentry{[40, 40]}
\end{axis}

\end{tikzpicture}}}}%
\end{minipage}%
\begin{minipage}{0.23\textwidth}
\subfigure[$\delta = 0.1$]{\label{fig:gridworld_eval_1}\scalebox{0.4}{\hspace{-1.0em}\hbox{\hspace{-1.0em}\input{GridWorld/gridworld_evaluation1_3162}}}}%
\end{minipage}
\begin{minipage}{0.23\textwidth}
\subfigure[$\delta=0.5$]{\label{fig:gridworld_eval_5}\scalebox{0.4}{\hbox{\hspace{-1.0em}\input{GridWorld/gridworld_evaluation5_3162}}}}
\end{minipage}
\begin{minipage}{0.23\textwidth}
\subfigure[$\delta=0.9$]{\label{fig:gridworld_eval_9}\scalebox{0.4}{\hbox{\hspace{-1.0em}\input{GridWorld/gridworld_evaluation9_3162}}}}
\end{minipage}
\caption{\textbf{Gridworld}. (a) We plot the absolute error between the transition ratios and the estimated transition ratios averaged across all state-action-state pairs. (b, c, d) Evaluation using our approach is depicted in green, importance sampling baseline in blue, the oracle in red and the true value.}
\label{fig: gridworld_2}
\vspace{-0.7cm}
\end{figure*}
In this experiment, we show that our algorithm is able to approximate the ratio of transition probabilites. We also show that it is also possible to validate the policy performance in the test environment using the simulator and our estimator. To that end, we define $\text{Gridworld}_\varepsilon$ as a $n \times n$ sized grid space. The agent starts with a fixed initial state at the bottom left corner of the grid and can take one of the four actions \{N, S, E, W\}. The MDP is designed such that whenever the agent takes an action, (e.g., N) they move in that direction with 1-$\varepsilon$ probability and with $\varepsilon$ probability end up in any of the remaining adjacent states. An episode ends when the agent reaches the goal state in the top-right corner of the grid. 

We use $\text{Gridworld}$ with ${\varepsilon = 0.3}$ as the training MDP and $\text{Gridworld}$ with ${\varepsilon = 0.1}$ as the testing MDP. In figure \ref{fig:gridworld_1}, we plot the mean absolute error of the transition probabilities over all the state-action-state pairs along with one standard deviation. The absolute error is plotted against the sample size which was used to train the estimator. We also compare our estimation error against three different grid sizes $\{10 \times 10, 20 \times 20, 40 \times 40\}$. We observe that the estimation error decreases with sample size, while larger grids require more samples to achieve reasonable error margins. These results highlight the dependence of the estimator on the size of the state space and the need for more samples in more complex environments.

Next, we evaluate the performance of a policy using the training environment and the estimated transition probability. Using equation \ref{eq: value_approx}, we evaluate performance for a range of policies (parameterised by $\delta$) over 200 timesteps. The functional form of these policies can be written as follows, $\pi_\delta(\cdot | s) = (1-\delta) \cdot U(\{N, S, E, W\}) + \delta \cdot \pi_E(.|s)$. $U(.)$ is the policy that picks each of the four actions randomly at every state and $\pi_E(.)$ is the expert policy. We evaluate the performance for $\delta=0.1, 0.5, 0.9$ over 200 timesteps respectively. These plots are presented in figures \ref{fig:gridworld_eval_1}, \ref{fig:gridworld_eval_5} and \ref{fig:gridworld_eval_9} respectively.

In each of these figures, we compare our performance against the true value which is the actual performance in the test environment, the oracle, and the importance sampling estimator. The oracle is the estimator that knows the true value of these transition probabilities, so the only error that arises in the evaluation is the error due to averaging over finite number of samples. We also compare our method against the importance sampling (IS) estimator. IS estimators trace their origins from Off Policy Evaluation methods where we have data from a behavior policy and we wish to evaluate performance for a given target policy. In all of these figures, we notice that the importance sampling method performs better whenever the target policy is close to behavior policy, the performance deteriorates as target policy moves away from the behavior policy. We observe that our method tends to generalize well over a the range of different policies. 

It is important to note that the best performance that our estimator can ideally get is the one approximated by the oracle. One might notice here that our estimator tends to perform better than the oracle. A possible reason could be that, while optimizing, we never constrain the neural network to learn over the structure of the ratio of two probability distributions. For example, suppose the function $\hat{\beta} \in \mathcal{G}$ is the optimum function that maximizes equation \ref{eq: optimizer}. On top of optimizing the equation, $\hat{\beta}$ also needs to satisfy the constraint 
\begin{align*}
\sum_{x} \hat{\beta}(x)q(x) = 1.    
\end{align*}
We assume that this constraint will be implicitly satisfied while optimizing, but since the algorithm empirical samples from distributions to calculate $\hat{\beta}$, it may not satisfy the above constraint.
\subsection{Cartpole Environment}
\begin{figure*}
\centering
\begin{minipage}{0.23\textwidth}
\subfigure[gravity = $7.5 m/s^2$]{\label{fig:cartpole_75}\scalebox{0.40}{\hbox{\hspace{-1.0em}
\begin{tikzpicture}

\definecolor{color0}{rgb}{0,0.266666666666667,0.105882352941176}
\definecolor{color1}{rgb}{0.0313725490196078,0.250980392156863,0.505882352941176}

\begin{axis}[
legend cell align={left},
legend style={fill opacity=0.8, draw opacity=1, text opacity=1, at={(1.0,0.8)}, anchor=east, draw=white!80!black},
tick align=outside,
tick pos=left,
x grid style={white!69.0196078431373!black},
xlabel={$\delta$},
xmajorgrids,
xmin=-0.05, xmax=1.05,
xtick style={color=black},
y grid style={white!69.0196078431373!black},
ylabel={Average Returns},
ymajorgrids,
ymin=0.156415544996506, ymax=0.92,
ytick style={color=black},
ytick={0.1,0.2,0.3,0.4,0.5,0.6,0.7},
yticklabels={0.1,0.2,0.3,0.4,0.5,0.6,0.7}
]
\path [draw=black, fill=black, opacity=0.1]
(axis cs:0,0.629501423509269)
--(axis cs:0,0.618683913963555)
--(axis cs:0.2,0.561965379643755)
--(axis cs:0.4,0.522118423279378)
--(axis cs:0.6,0.384923130554098)
--(axis cs:0.8,0.298875548204678)
--(axis cs:1,0.179330079311668)
--(axis cs:1,0.198485805848516)
--(axis cs:1,0.198485805848516)
--(axis cs:0.8,0.328617631220427)
--(axis cs:0.6,0.420833791277578)
--(axis cs:0.4,0.54791764414569)
--(axis cs:0.2,0.582626298026894)
--(axis cs:0,0.629501423509269)
--cycle;

\path [draw=color0, fill=color0, opacity=0.1]
(axis cs:0,0.604591551706688)
--(axis cs:0,0.587398812247486)
--(axis cs:0.2,0.498193948788693)
--(axis cs:0.4,0.470227820550075)
--(axis cs:0.6,0.365933062169921)
--(axis cs:0.8,0.286383111240799)
--(axis cs:1,0.200625736225378)
--(axis cs:1,0.2170831212885)
--(axis cs:1,0.2170831212885)
--(axis cs:0.8,0.317368592974749)
--(axis cs:0.6,0.40053280294972)
--(axis cs:0.4,0.505950691427166)
--(axis cs:0.2,0.53244998411257)
--(axis cs:0,0.604591551706688)
--cycle;

\path [draw=red!49.8039215686275!black, fill=red!49.8039215686275!black, opacity=0.1]
(axis cs:0,0.633967658726771)
--(axis cs:0,0.633967658726771)
--(axis cs:0.2,0.633967658726771)
--(axis cs:0.4,0.633967658726771)
--(axis cs:0.6,0.633967658726771)
--(axis cs:0.8,0.633967658726771)
--(axis cs:1,0.633967658726771)
--(axis cs:1,0.633967658726771)
--(axis cs:1,0.633967658726771)
--(axis cs:0.8,0.633967658726771)
--(axis cs:0.6,0.633967658726771)
--(axis cs:0.4,0.633967658726771)
--(axis cs:0.2,0.633967658726771)
--(axis cs:0,0.633967658726771)
--cycle;

\addplot [ultra thick, black, dashed]
table {%
0 0.624092668736412
0.2 0.572295838835325
0.4 0.535018033712534
0.6 0.402878460915838
0.8 0.313746589712552
1 0.188907942580092
};
\addlegendentry{True Value}
\addplot [ultra thick, color0]
table {%
0 0.595995181977087
0.2 0.515321966450632
0.4 0.48808925598862
0.6 0.38323293255982
0.8 0.301875852107774
1 0.208854428756939
};
\addlegendentry{OEE (ours)}
\addplot [ultra thick, red!49.8039215686275!black]
table {%
0 0.633967658726771
0.2 0.633967658726771
0.4 0.633967658726771
0.6 0.633967658726771
0.8 0.633967658726771
1 0.633967658726771
};
\addlegendentry{MLE}
\addplot [ultra thick, color1]
table {%
0.2 0.2367942844232
0.4 0.630281835901953
0.6 0.628707744294357
0.8 0.637423381853606
1 0.637620765614916
};
\addlegendentry{IS}
\end{axis}

\end{tikzpicture}}}}
\end{minipage}%
\begin{minipage}{0.23\textwidth}
\subfigure[gravity = $10.0 m/s^2$]{\label{fig:cartpole_100}\scalebox{0.40}{\hbox{\hspace{-1.0em}
\begin{tikzpicture}

\definecolor{color0}{rgb}{0,0.266666666666667,0.105882352941176}
\definecolor{color1}{rgb}{0.0313725490196078,0.250980392156863,0.505882352941176}

\begin{axis}[
legend cell align={left},
legend style={fill opacity=0.8, draw opacity=1, text opacity=1, at={(1.0,0.8)}, anchor=east, draw=white!80!black},
tick align=outside,
tick pos=left,
x grid style={white!69.0196078431373!black},
xlabel={$\delta$},
xmajorgrids,
xmin=-0.05, xmax=1.05,
xtick style={color=black},
y grid style={white!69.0196078431373!black},
ylabel={Average Returns},
ymajorgrids,
ymin=0.164828027187512, ymax=0.95,
ytick style={color=black},
ytick={0.1,0.2,0.3,0.4,0.5,0.6,0.7},
yticklabels={0.1,0.2,0.3,0.4,0.5,0.6,0.7}
]
\path [draw=black, fill=black, opacity=0.1]
(axis cs:0,0.623472849561475)
--(axis cs:0,0.610495606158438)
--(axis cs:0.2,0.604352286203161)
--(axis cs:0.4,0.477567832275723)
--(axis cs:0.6,0.30430530471784)
--(axis cs:0.8,0.236749655601927)
--(axis cs:1,0.196495989777101)
--(axis cs:1,0.22041808312982)
--(axis cs:1,0.22041808312982)
--(axis cs:0.8,0.260182866580816)
--(axis cs:0.6,0.331577811055625)
--(axis cs:0.4,0.505756560459993)
--(axis cs:0.2,0.619071571623247)
--(axis cs:0,0.623472849561475)
--cycle;

\path [draw=color0, fill=color0, opacity=0.1]
(axis cs:0,0.61032372703168)
--(axis cs:0,0.607624558007213)
--(axis cs:0.2,0.580007489652134)
--(axis cs:0.4,0.459236642326026)
--(axis cs:0.6,0.360135538959402)
--(axis cs:0.8,0.276360047085237)
--(axis cs:1,0.187295902860897)
--(axis cs:1,0.190486145295194)
--(axis cs:1,0.190486145295194)
--(axis cs:0.8,0.282762811281139)
--(axis cs:0.6,0.368751902900596)
--(axis cs:0.4,0.468716198475053)
--(axis cs:0.2,0.583243496003518)
--(axis cs:0,0.61032372703168)
--cycle;

\path [draw=red!49.8039215686275!black, fill=red!49.8039215686275!black, opacity=0.1]
(axis cs:0,0.633967658726771)
--(axis cs:0,0.633967658726771)
--(axis cs:0.2,0.633967658726771)
--(axis cs:0.4,0.607122280419392)
--(axis cs:0.6,0.633967658726771)
--(axis cs:0.8,0.633967658726771)
--(axis cs:1,0.633967658726771)
--(axis cs:1,0.633967658726771)
--(axis cs:1,0.633967658726771)
--(axis cs:0.8,0.633967658726771)
--(axis cs:0.6,0.633967658726771)
--(axis cs:0.4,0.626122868337956)
--(axis cs:0.2,0.633967658726771)
--(axis cs:0,0.633967658726771)
--cycle;

\addplot [ultra thick, black, dashed]
table {%
0 0.616984227859957
0.2 0.611711928913204
0.4 0.491662196367858
0.6 0.317941557886733
0.8 0.248466261091372
1 0.20845703645346
};
\addlegendentry{True Value}
\addplot [ultra thick, color0]
table {%
0 0.608974142519446
0.2 0.581625492827826
0.4 0.463976420400539
0.6 0.364443720929999
0.8 0.279561429183188
1 0.188891024078046
};
\addlegendentry{OEE (ours)}
\addplot [ultra thick, red!49.8039215686275!black]
table {%
0 0.633967658726771
0.2 0.633967658726771
0.4 0.616622574378674
0.6 0.633967658726771
0.8 0.633967658726771
1 0.633967658726771
};
\addlegendentry{MLE}
\addplot [ultra thick, color1]
table {%
0.2 0.264192958086922
0.4 0.629774110691547
0.6 0.632703310439804
0.8 0.636525854096907
1 0.636653416328594
};
\addlegendentry{IS}
\end{axis}

\end{tikzpicture}}}}
\end{minipage}%
\begin{minipage}{0.23\textwidth}
\subfigure[gravity = $12.5 m/s^2$]{\label{fig:cartpole_125}\scalebox{0.40}{\hbox{\hspace{-1.0em}
\begin{tikzpicture}

\definecolor{color0}{rgb}{0,0.266666666666667,0.105882352941176}
\definecolor{color1}{rgb}{0.0313725490196078,0.250980392156863,0.505882352941176}

\begin{axis}[
legend cell align={left},
legend style={fill opacity=0.8, draw opacity=1, text opacity=1, at={(1.0,0.8)}, anchor=east, draw=white!80!black},
tick align=outside,
tick pos=left,
x grid style={white!69.0196078431373!black},
xlabel={$\delta$},
xmajorgrids,
xmin=-0.05, xmax=1.05,
xtick style={color=black},
y grid style={white!69.0196078431373!black},
ylabel={Average Returns},
ymajorgrids,
ymin=0.18361255853611, ymax=0.92,
ytick style={color=black},
ytick={0.1,0.2,0.3,0.4,0.5,0.6,0.7},
yticklabels={0.1,0.2,0.3,0.4,0.5,0.6,0.7}
]
\path [draw=black, fill=black, opacity=0.1]
(axis cs:0,0.624120886341571)
--(axis cs:0,0.612636619922068)
--(axis cs:0.2,0.577891321374915)
--(axis cs:0.4,0.436312133877623)
--(axis cs:0.6,0.375276271434724)
--(axis cs:0.8,0.260643441634365)
--(axis cs:1,0.205231856816092)
--(axis cs:1,0.221778591403141)
--(axis cs:1,0.221778591403141)
--(axis cs:0.8,0.283753768031099)
--(axis cs:0.6,0.405399425731694)
--(axis cs:0.4,0.468252208748827)
--(axis cs:0.2,0.599294032867457)
--(axis cs:0,0.624120886341571)
--cycle;

\path [draw=color0, fill=color0, opacity=0.1]
(axis cs:0,0.597801671760834)
--(axis cs:0,0.578545653862536)
--(axis cs:0.2,0.59625100950277)
--(axis cs:0.4,0.549415577023681)
--(axis cs:0.6,0.390247467855842)
--(axis cs:0.8,0.31150406430007)
--(axis cs:1,0.226116821219386)
--(axis cs:1,0.241651327152207)
--(axis cs:1,0.241651327152207)
--(axis cs:0.8,0.337441582461229)
--(axis cs:0.6,0.415664923471333)
--(axis cs:0.4,0.568110451361023)
--(axis cs:0.2,0.608969051347696)
--(axis cs:0,0.597801671760834)
--cycle;

\path [draw=red!49.8039215686275!black, fill=red!49.8039215686275!black, opacity=0.1]
(axis cs:0,0.633967658726771)
--(axis cs:0,0.633967658726771)
--(axis cs:0.2,0.633967658726771)
--(axis cs:0.4,0.631331962602969)
--(axis cs:0.6,0.633967658726771)
--(axis cs:0.8,0.633967658726771)
--(axis cs:1,0.633967658726771)
--(axis cs:1,0.633967658726771)
--(axis cs:1,0.633967658726771)
--(axis cs:0.8,0.633967658726771)
--(axis cs:0.6,0.633967658726771)
--(axis cs:0.4,0.633197452273239)
--(axis cs:0.2,0.633967658726771)
--(axis cs:0,0.633967658726771)
--cycle;

\addplot [ultra thick, black, dashed]
table {%
0 0.61837875313182
0.2 0.588592677121186
0.4 0.452282171313225
0.6 0.390337848583209
0.8 0.272198604832732
1 0.213505224109617
};
\addlegendentry{True Value}
\addplot [ultra thick, color0]
table {%
0 0.588173662811685
0.2 0.602610030425233
0.4 0.558763014192352
0.6 0.402956195663588
0.8 0.32447282338065
1 0.233884074185796
};
\addlegendentry{OEE (ours)}
\addplot [ultra thick, red!49.8039215686275!black]
table {%
0 0.633967658726771
0.2 0.633967658726771
0.4 0.632264707438104
0.6 0.633967658726771
0.8 0.633967658726771
1 0.633967658726771
};
\addlegendentry{MLE}
\addplot [ultra thick, color1]
table {%
0.2 0.257681596974562
0.4 0.62469682808488
0.6 0.636265165812878
0.8 0.636889740171387
1 0.637617822415745
};
\addlegendentry{IS}
\end{axis}

\end{tikzpicture}}}}
\end{minipage}%
\begin{minipage}{0.23\textwidth}
\subfigure[gravity = $15.0 m/s^2$]{\label{fig:cartpole_125}\scalebox{0.40}{\hbox{\hspace{-1.0em}
\begin{tikzpicture}

\definecolor{color0}{rgb}{0,0.266666666666667,0.105882352941176}
\definecolor{color1}{rgb}{0.0313725490196078,0.250980392156863,0.505882352941176}

\begin{axis}[
legend cell align={left},
legend style={fill opacity=0.8, draw opacity=1, text opacity=1, at={(1.0,0.8)}, anchor=east, draw=white!80!black},
tick align=outside,
tick pos=left,
x grid style={white!69.0196078431373!black},
xlabel={$\delta$},
xmajorgrids,
xmin=-0.05, xmax=1.05,
xtick style={color=black},
y grid style={white!69.0196078431373!black},
ylabel={Average Returns},
ymajorgrids,
ymin=0.167278384827473, ymax=0.92,
ytick style={color=black},
ytick={0.1,0.2,0.3,0.4,0.5,0.6,0.7},
yticklabels={0.1,0.2,0.3,0.4,0.5,0.6,0.7}
]
\path [draw=black, fill=black, opacity=0.1]
(axis cs:0,0.619071384886136)
--(axis cs:0,0.604404584236168)
--(axis cs:0.2,0.575817265488857)
--(axis cs:0.4,0.412821676588405)
--(axis cs:0.6,0.305772429277514)
--(axis cs:0.8,0.217082794985583)
--(axis cs:1,0.20202114034859)
--(axis cs:1,0.214082979803414)
--(axis cs:1,0.214082979803414)
--(axis cs:0.8,0.239487674061108)
--(axis cs:0.6,0.326415463656888)
--(axis cs:0.4,0.446634104901897)
--(axis cs:0.2,0.592570128900893)
--(axis cs:0,0.619071384886136)
--cycle;

\path [draw=color0, fill=color0, opacity=0.1]
(axis cs:0,0.578329867696302)
--(axis cs:0,0.553897086845757)
--(axis cs:0.2,0.481838631371923)
--(axis cs:0.4,0.38924361711064)
--(axis cs:0.6,0.293433822484969)
--(axis cs:0.8,0.243965615949782)
--(axis cs:1,0.189667561754795)
--(axis cs:1,0.20086742222702)
--(axis cs:1,0.20086742222702)
--(axis cs:0.8,0.2577641478123)
--(axis cs:0.6,0.312060514050609)
--(axis cs:0.4,0.41780803843611)
--(axis cs:0.2,0.513358838976962)
--(axis cs:0,0.578329867696302)
--cycle;

\path [draw=red!49.8039215686275!black, fill=red!49.8039215686275!black, opacity=0.1]
(axis cs:0,0.633967658726771)
--(axis cs:0,0.633967658726771)
--(axis cs:0.2,0.633967658726771)
--(axis cs:0.4,0.633967658726771)
--(axis cs:0.6,0.633967658726771)
--(axis cs:0.8,0.633967658726771)
--(axis cs:1,0.633967658726771)
--(axis cs:1,0.633967658726771)
--(axis cs:1,0.633967658726771)
--(axis cs:0.8,0.633967658726771)
--(axis cs:0.6,0.633967658726771)
--(axis cs:0.4,0.633967658726771)
--(axis cs:0.2,0.633967658726771)
--(axis cs:0,0.633967658726771)
--cycle;

\addplot [ultra thick, black, dashed]
table {%
0 0.611737984561152
0.2 0.584193697194875
0.4 0.429727890745151
0.6 0.316093946467201
0.8 0.228285234523345
1 0.208052060076002
};
\addlegendentry{True Value}
\addplot [ultra thick, color0]
table {%
0 0.566113477271029
0.2 0.497598735174442
0.4 0.403525827773375
0.6 0.302747168267789
0.8 0.250864881881041
1 0.195267491990908
};
\addlegendentry{OEE (ours)}
\addplot [ultra thick, red!49.8039215686275!black]
table {%
0 0.633967658726771
0.2 0.633967658726771
0.4 0.633967658726771
0.6 0.633967658726771
0.8 0.633967658726771
1 0.633967658726771
};
\addlegendentry{MLE}
\addplot [ultra thick, color1]
table {%
0.2 0.256370648755857
0.4 0.625565364018606
0.6 0.637062822654234
0.8 0.636819457913452
1 0.637451100301224
};
\addlegendentry{IS}
\end{axis}

\end{tikzpicture}}}}
\end{minipage}
\caption{\textbf{Cartpole Environment}. Evaluation using our algorithm is depicted in green, MLE baseline is in red, the importance sampling baseline is in blue, and the true performance is in black.}
\label{fig: cartpole_oee}
\end{figure*}
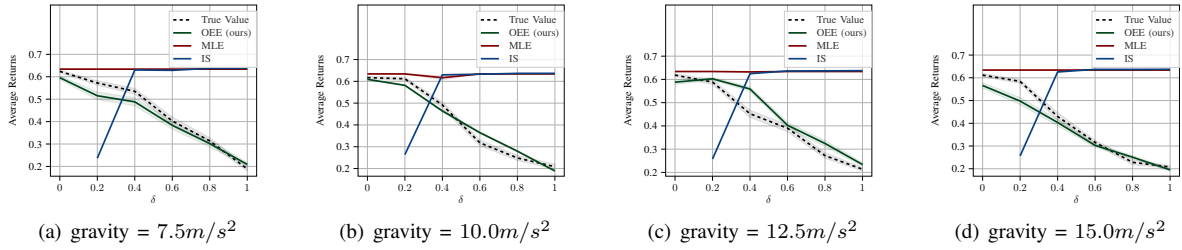

We now examine the Cartpole environment, which has a continuous state space and discrete action space. Since the Carpole environment is a highly unstable environments, good sampling strategies are very important, as the estimator tends to fails to generalize if only poor data is collected.
For the Cartpole environment, the training MDP has gravity $g = 10.0m/s^2$. We evaluate our performance on a range of test MDPs with gravity $g = \{7.5, 10.0, 12.5, 15.0\} m/s^2$.
As in the gridworld environment, we demonstrate our performance over a range of policies parameterized by $\delta$, $\pi_\delta(. | s) = \delta \cdot U(|\mathcal{A}|) + (1-\delta) \cdot \pi_E(.|s)$. The expert policy is a deterministic policy trained using the cross entropy method~\cite{DBLP:conf/icml/MannorRG03}. We plot of our results for each of the target environments in figure \ref{fig: cartpole_oee} over 100 timesteps.  We compare our results against a model based baseline which learns the model using the same data that was used to learn the transition ratio. We also use an IS based OPE as a baseline to compare performance. 

We observe that our algorithm closely approximates the true performance over a range of policies in the target environment. We also observe that as divergence between the two environments increases, the performance predicted by our algorithm tends to deteriorate. In comparison, the importance sampling baseline works only well when the target policies near the data collection policy but the performance soon deteriorates after that. On the other hand, the MLE which learns the model of the environment using the target environment data, tends to perform worse.
\subsection{Reacher Environment}
\begin{figure}
\begin{flushleft}
\begin{minipage}{0.23\textwidth}
\subfigure[$l_{te} = 2 l_{tr} $]{\label{fig:reacher_2}\scalebox{0.4}{\hbox{\hspace{-1.0em}
\begin{tikzpicture}

\definecolor{color0}{rgb}{0,0.266666666666667,0.105882352941176}
\definecolor{color1}{rgb}{0.0313725490196078,0.250980392156863,0.505882352941176}

\begin{axis}[
legend cell align={left},
legend style={fill opacity=0.8, draw opacity=1, text opacity=1, at={(1.0,0.84)}, anchor=east, draw=white!80!black},
tick align=outside,
tick pos=left,
x grid style={white!69.0196078431373!black},
xlabel={$\delta$},
xmajorgrids,
xmin=0.06, xmax=0.94,
xtick style={color=black},
y grid style={white!69.0196078431373!black},
ylabel={Average Returns},
ymajorgrids,
ymin=-4.37343414856659, ymax=36,
ytick style={color=black}
]
\path [draw=black, fill=black, opacity=0.1]
(axis cs:0.1,9.71017326267829)
--(axis cs:0.1,7.38088083276743)
--(axis cs:0.3,4.10010782929016)
--(axis cs:0.5,2.4256962375125)
--(axis cs:0.7,4.27919759943841)
--(axis cs:0.9,0.80296589113128)
--(axis cs:0.9,3.06518574476063)
--(axis cs:0.9,3.06518574476063)
--(axis cs:0.7,6.58564511741213)
--(axis cs:0.5,4.68880258438846)
--(axis cs:0.3,6.37944066803125)
--(axis cs:0.1,9.71017326267829)
--cycle;

\path [draw=color0, fill=color0, opacity=0.1]
(axis cs:0.1,10.0329923418342)
--(axis cs:0.1,7.89126144219857)
--(axis cs:0.3,6.25458121084373)
--(axis cs:0.5,4.35977141710789)
--(axis cs:0.7,0.665267442620762)
--(axis cs:0.9,0.155957687806908)
--(axis cs:0.9,1.14793879370436)
--(axis cs:0.9,1.14793879370436)
--(axis cs:0.7,1.69186587456037)
--(axis cs:0.5,6.00965256253915)
--(axis cs:0.3,7.44644196303983)
--(axis cs:0.1,10.0329923418342)
--cycle;

\path [draw=red!49.8039215686275!black, fill=red!49.8039215686275!black, opacity=0.1]
(axis cs:0.1,2.83098552615828)
--(axis cs:0.1,1.2751680829877)
--(axis cs:0.3,-0.15038189335427)
--(axis cs:0.5,0.260166775972344)
--(axis cs:0.7,-2.09372580456982)
--(axis cs:0.9,-3.04199083923372)
--(axis cs:0.9,-1.38291245948429)
--(axis cs:0.9,-1.38291245948429)
--(axis cs:0.7,-0.437678586564373)
--(axis cs:0.5,1.82968469402705)
--(axis cs:0.3,1.52551883691006)
--(axis cs:0.1,2.83098552615828)
--cycle;

\path [draw=color1, fill=color1, opacity=0.1]
(axis cs:0.1,21.2049160173137)
--(axis cs:0.1,18.2128068883449)
--(axis cs:0.3,20.0803265225469)
--(axis cs:0.5,20.1113699407643)
--(axis cs:0.7,20.2233324126766)
--(axis cs:0.9,16.6874412750815)
--(axis cs:0.9,19.898388206697)
--(axis cs:0.9,19.898388206697)
--(axis cs:0.7,23.5358769897854)
--(axis cs:0.5,23.5868753474236)
--(axis cs:0.3,23.2027126147284)
--(axis cs:0.1,21.2049160173137)
--cycle;

\addplot [ultra thick, black, dashed]
table {%
0.1 8.54552704772286
0.3 5.23977424866071
0.5 3.55724941095048
0.7 5.43242135842527
0.9 1.93407581794596
};
\addlegendentry{True Value}
\addplot [ultra thick, color0]
table {%
0.1 8.96212689201637
0.3 6.85051158694178
0.5 5.18471198982352
0.7 1.17856665859057
0.9 0.651948240755634
};
\addlegendentry{OEE (ours)}
\addplot [ultra thick, red!49.8039215686275!black]
table {%
0.1 2.05307680457299
0.3 0.687568471777897
0.5 1.0449257349997
0.7 -1.2657021955671
0.9 -2.21245164935901
};
\addlegendentry{Simulated}
\addplot [ultra thick, color1]
table {%
0.1 19.7088614528293
0.3 21.6415195686377
0.5 21.849122644094
0.7 21.879604701231
0.9 18.2929147408893
};
\addlegendentry{MLE}
\end{axis}

\end{tikzpicture}}}}%
\end{minipage}
\begin{minipage}{0.23\textwidth}
\subfigure[$l_{te} = 0.5 l_{tr} $]{\label{fig:reacher_5}\scalebox{0.4}{\hbox{\hspace{-1.0em}
\begin{tikzpicture}

\definecolor{color0}{rgb}{0,0.266666666666667,0.105882352941176}
\definecolor{color1}{rgb}{0.0313725490196078,0.250980392156863,0.505882352941176}

\begin{axis}[
legend cell align={left},
legend style={fill opacity=0.8, draw opacity=1, text opacity=1, at={(1.0,0.84)},anchor=east, draw=white!80!black},
tick align=outside,
tick pos=left,
x grid style={white!69.0196078431373!black},
xlabel={$\delta$},
xmajorgrids,
xmin=0.06, xmax=0.94,
xtick style={color=black},
y grid style={white!69.0196078431373!black},
ylabel={Average Returns},
ymajorgrids,
ymin=-11.0643959485254, ymax=11.0,
ytick style={color=black}
]
\path [draw=black, fill=black, opacity=0.1]
(axis cs:0.1,-1.2875140616707)
--(axis cs:0.1,-2.43066095429405)
--(axis cs:0.3,-4.40012753347477)
--(axis cs:0.5,-5.49024774612798)
--(axis cs:0.7,-7.4887615082334)
--(axis cs:0.9,-10.1278369042291)
--(axis cs:0.9,-9.13233497447182)
--(axis cs:0.9,-9.13233497447182)
--(axis cs:0.7,-6.35041385065918)
--(axis cs:0.5,-4.40991266310746)
--(axis cs:0.3,-3.45040275101928)
--(axis cs:0.1,-1.2875140616707)
--cycle;

\path [draw=color0, fill=color0, opacity=0.1]
(axis cs:0.1,-1.5807993033975)
--(axis cs:0.1,-2.39490354354678)
--(axis cs:0.3,-1.95619510601844)
--(axis cs:0.5,-3.74730528939425)
--(axis cs:0.7,-4.3562470526558)
--(axis cs:0.9,-7.57016957137322)
--(axis cs:0.9,-6.48082547109842)
--(axis cs:0.9,-6.48082547109842)
--(axis cs:0.7,-2.98680165121988)
--(axis cs:0.5,-1.93711563290712)
--(axis cs:0.3,-1.01624104579214)
--(axis cs:0.1,-1.5807993033975)
--cycle;

\path [draw=red!49.8039215686275!black, fill=red!49.8039215686275!black, opacity=0.1]
(axis cs:0.1,2.31641806988832)
--(axis cs:0.1,0.719725948814063)
--(axis cs:0.3,0.109097419632275)
--(axis cs:0.5,-0.593610513461785)
--(axis cs:0.7,-2.80596666231019)
--(axis cs:0.9,-3.91035736110178)
--(axis cs:0.9,-2.05694006181589)
--(axis cs:0.9,-2.05694006181589)
--(axis cs:0.7,-1.17064215227282)
--(axis cs:0.5,1.0202017146126)
--(axis cs:0.3,1.87832630000865)
--(axis cs:0.1,2.31641806988832)
--cycle;

\path [draw=color1, fill=color1, opacity=0.1]
(axis cs:0.1,8.00623249190011)
--(axis cs:0.1,6.39915144830007)
--(axis cs:0.3,6.80096417541834)
--(axis cs:0.5,4.51300132829493)
--(axis cs:0.7,-0.405727243641089)
--(axis cs:0.9,-5.43610528024188)
--(axis cs:0.9,-3.43223510845331)
--(axis cs:0.9,-3.43223510845331)
--(axis cs:0.7,1.28632099880165)
--(axis cs:0.5,6.29686660497127)
--(axis cs:0.3,8.60334398169657)
--(axis cs:0.1,8.00623249190011)
--cycle;

\addplot [ultra thick, black, dashed]
table {%
0.1 -1.85908750798237
0.3 -3.92526514224703
0.5 -4.95008020461772
0.7 -6.91958767944629
0.9 -9.63008593935045
};
\addlegendentry{True Value}
\addplot [ultra thick, color0]
table {%
0.1 -1.98785142347214
0.3 -1.48621807590529
0.5 -2.84221046115068
0.7 -3.67152435193784
0.9 -7.02549752123582
};
\addlegendentry{OEE (ours)}
\addplot [ultra thick, red!49.8039215686275!black]
table {%
0.1 1.51807200935119
0.3 0.993711859820461
0.5 0.213295600575407
0.7 -1.98830440729151
0.9 -2.98364871145883
};
\addlegendentry{Simulated}
\addplot [ultra thick, color1]
table {%
0.1 7.20269197010009
0.3 7.70215407855745
0.5 5.4049339666331
0.7 0.440296877580281
0.9 -4.4341701943476
};
\addlegendentry{MLE}
\end{axis}

\end{tikzpicture}}}}
\end{minipage}
\caption{\textbf{Reacher Environment}. Evaluation using our algorithm is depicted in green, the simulator performance in red, the MLE baseline is in blue, and the true value is in black.}
\vspace{-0.7cm}
\label{fig: oee_reacher}
\end{flushleft}
\end{figure}
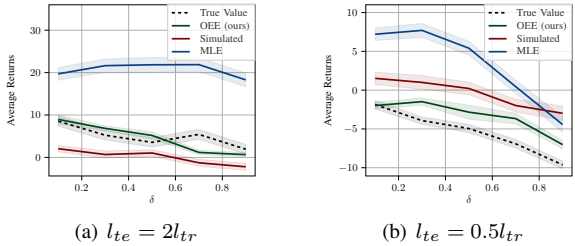
We also evaluate our performance on a continuous state, continous action Reacher Environment. For these experiments pertaining to the Reacher environment, the training environment has both link lengths set to $0.1 m$. We validate the performance of our algorithm on a set of two test environments. The first target environment has twice the length of the link as the training environment. The second target environment, has half the length of the link in the training environment. 

We evaluate the performance using the training environment and the OEE estimator over a range of policies parameterized by $\epsilon$, $\pi_\delta(. | s) = \delta \cdot U(|\mathcal{A}|) + (1-\delta) \cdot \pi_E(.|s)$. These performance is reported along with their standard error in figure \ref{fig: oee_reacher}. For comparison we also reported performances predicted by the simulator, and the one predicted by a learned model that trained over the same data collected over the target environment. We observe that even when the difference in the performance between the training and the testing environment is large, our method is able to make up for that difference and predict performance accurately over a range of policies.  
\subsection{Real World Kinova Environment}
\begin{table}[!htb]
\captionsetup{singlelinecheck = false, justification=justified}
\captionof{table}{Performance as predicted on a Kinova Robot}
\normalsize
\label{tab:kinova}
\begin{center} 
\begin{tabular}{@{}lcc@{}}
\toprule 
Algorithm & Mean & Std\\
\midrule
True Value & -0.911 & 0.016 \\
OEE & -1.095 & 0.101 \\
Simulator & -1.181 & 0.170 \\
\bottomrule
\end{tabular}
\end{center}
\vspace{-0.5cm}
\end{table}
To demonstrate the practicality of our algorithm in a real world setting Sim2Real tasks, we evaluate the performance of our algorithm on 
Kinova Gen3 modular robotic arm. We use a  Kinova simulator that was designed for training RL algorithms~\cite{chang2020robot}. Using this simulator along with demonstration data collected on the Robot, we optimize equation \ref{eq: optimizer} calculate the ratio of transition probabilities. 

Using this estimated conditional probability we evaluate the performance of a proportional controller that maneuvers the robot arm from an initial point to a final goal state. To that end, we choose 5 different goal states and the cost function incurred at each of the goal states is averaged over 10 runs each. We compare the performance against the one predicted by our algorithm using the simulator and the estimated conditional probability over 50 trajectories. In addition, we also report the returns approximated by the simulator. All of these results are tabulated in Table \ref{tab:kinova}. 

We observe that the performance of the simulator as compared to the one observed on the robot is off by around 30\%. Using our approach we are able to decrease that discrepancy to around 20\%. It is important to read these results in context. We note performance improvement is important because it decreases the Sim2Real gap, which helps us learn policies that are optimal in the target environment 
\section{Conclusion and Remarks} \label{sec: conclude}
In this paper, we propose an algorithm that can evaluate performance of a deployed agent using a minimal amount of data in the target environment along with access to the training environment. We first estimate the transition probability ratio between two MDPs using convex risk minimization algorithm. We demonstrate that this ratio can be used to measure performance of an RL policy in the target environment using the training environment and the estimator. 

Our experiments demonstrate that, using data from the real world which combined with the simulator, one can indeed approximate performance of any policy. We believe that the ability to validate performance of agents in another environment can further help us learn policies in the training environment that are well performing in the target domain. However, we note that the quality of data collected over target environment does impact performance. It was observed that data collected using a sub-optimal policy tends to generalize well. While we demonstrated our method on standard simulators plus real world robot, exploring the utility of such methods on more complex systems will be explored in future work. 

\newpage
\bibliography{references}
\bibliographystyle{IEEEtran}
\end{document}